\documentclass{article}

\usepackage{PRIMEarxiv}

\usepackage[utf8]{inputenc} 
\usepackage[T1]{fontenc}    
\usepackage{hyperref}       
\usepackage{url}            
\usepackage{booktabs}       
\usepackage{amsfonts}       
\usepackage{nicefrac}       
\usepackage{microtype}      
\usepackage{lipsum}
\usepackage{fancyhdr}       
\usepackage{graphicx}
\usepackage{amsmath}
\usepackage{amsfonts}
\usepackage{amssymb}
\usepackage{graphicx}
\usepackage{appendix}
\usepackage [english]{babel}
\usepackage [autostyle, english = american]{csquotes}
\usepackage{booktabs}
\usepackage{multirow}
\MakeOuterQuote{"}
\usepackage{setspace}
\graphicspath{{images/}}
\usepackage{amsthm}
\usepackage{array}
\usepackage{cite}

\hypersetup{hidelinks}
\usepackage{longtable}
\usepackage{fancyhdr}
\usepackage{titletoc}
\usepackage{subcaption}
\usepackage{tabularx} 
 
\usepackage{float}
\usepackage{tablefootnote}
\usepackage{caption}
\usepackage{multirow}
\usepackage{xcolor}
\usepackage{textcomp}

\newtheorem{lemma}{Lemma}
\graphicspath{{media/}}

\title{Hyperparameter-Free Neurochaos Learning Algorithm for Classification}

\author{
  Akhila Henry \\
  Department of Mathematics\\
  Amrita Vishwa Vidyapeetham\\
  Amritapuri Campus \\
  Kollam, 690525, Kerala, India.\\
  \texttt{akhilahenryu@am.amrita.edu} \\
   \And
  Nithin Nagaraj\\
  Complex Systems Programme\\ National Institute of Advanced Studies\\ Indian Institute of Science Campus\\ Bengaluru, 560012,
Karnataka, India.\\
  \texttt{nithin@nias.res.in} \\
}

\begin{document}
\maketitle

\begin{abstract}
 Neurochaos Learning (NL) is a brain-inspired classification framework that employs chaotic dynamics to extract features from input data and yields state-of-the-art performance on classification tasks. However, NL requires the tuning of multiple hyperparameters and computing of four chaotic features per input sample. In this paper, we propose {\it AutochaosNet} -- a novel, hyperparameter-free variant of the NL algorithm that eliminates the need for both training and parameter optimization. AutochaosNet leverages a universal chaotic sequence derived from the Champernowne constant and uses the input stimulus to define firing time bounds for feature extraction. Two simplified variants—TM AutochaosNet and TM-FR AutochaosNet—are evaluated against existing NL architecture - ChaosNet. Our results demonstrate that AutochaosNet achieves competitive or superior classification performance while significantly reducing training time due to reduced computational effort. In addition to eliminating training and hyperparameter tuning, AutochaosNet exhibits excellent generalisation capabilities, making it a scalable and efficient choice for real-world classification tasks. Future work will focus on identifying universal orbits under various chaotic maps and incorporating them into the NL framework to further enhance performance.
\end{abstract}


\section{Introduction}

The current trajectory of artificial intelligence research is geared toward enabling machines to learn autonomously. A wide variety of machine learning algorithms are being developed at an accelerated pace. Among these, brain-inspired learning algorithms have recently garnered significant attention. One such approach is the \textit{Neurochaos Learning (NL)} algorithm~\cite{chaosnet}, which draws inspiration from the dynamical behaviour of the human brain, particularly the presence of chaos in neural activity.

In $2023$, researchers introduced two distinct architectures under the NL framework: ChaosNet and CFX (Chaotic Feature Extraction) + ML~\cite{sethi2023neurochaos}. These architectures employ chaos through the use of a one-dimensional chaotic map known as the Skew Tent Map. For each dataset, a neural trace is computed, which terminates when it falls within a predefined noise threshold.

Given a dataset with $m$ samples and $n$ features, the ChaosNet and CFX+ML algorithms extract four chaotic features - Firing Time, Firing Rate, Energy, and Entropy - for each input feature, resulting in a transformed feature set of size $4n$. These features are subsequently fed into classical machine learning classifiers or used directly with cosine similarity for classification. The algorithm involves hyperparameters such as the initial condition and skew value of the chaotic map, as well as the noise level, all of which require tuning. This process increases the computational cost and complexity of feature extraction.

In this study, we introduce a novel variant of the Neurochaos Learning algorithm that completely eliminates the need for hyperparameter tuning by leveraging a recently proposed universal chaotic orbit~\cite{henry2025universal}, derived using the Decimal Shift Map. This approach aims to streamline the feature extraction process while maintaining the effectiveness of chaotic representations for learning tasks.

\section{Neurochaos Learning using Universal Orbit} 

Neurochaos Learning (NL) is an interdisciplinary approach that integrates concepts from neuroscience, chaos theory, and machine learning. A recent advancement in this field involves the introduction of a novel class of chaotic orbits known as universal orbits, specifically defined with respect to the Decimal Shift Map (DSM)~\cite{strogatz2018nonlinear} and the Continued Fraction Map~\cite{henry2025universal}. In this study, we employ the universal orbit under the DSM for chaotic feature extraction.

The orbit of the DSM is generated by successively shifting the digits of the initial point. Universal orbits under the DSM can be generated from uncountably many initial points. Among these, we focus on the orbit derived from the Champernowne constant, denoted by $$C=0.1234567891011\ldots498499,$$ truncated after the digits corresponding to the number 499. Also, this constant is known to be normal~\cite{normality} in base 10, ensuring that all finite digit sequences appear with equal frequency~\cite{henry2025universal}. The following lemma provides the theoretical basis for the occurrence of any natural number within this constant. 

\begin{lemma}\label{cmp_position}
    Let $c=0.1\text{ }2\text{ }3\text{ }4\text{ }5\text{ }6\text{ }7\text{ }8\text{ }9\text{ }10\text{ }11\ldots N(N+1)(N+2)\ldots$ be the Champernowne constant and $c_N=0.1234567891011\ldots (N-2)(N-1)$ be the truncated portion of $c$, where $N=a_1a_2\ldots a_d$, $d\geq 2$. Then the number of digits after the decimal point in $c_N$ is given by\\
    \begin{equation}\label{formula champ}
        {dN-(10+10^2+\ldots+10^{d-1})-1}.
    \end{equation}\\
    In other words, the number $N$ will occur in the decimal expansion of $c$ after $${dN-(10+10^2+\ldots+10^{d-1})-1}$$ digits from the decimal point.~\cite{henry2025universal}.
\end{lemma}
A detailed proof of the lemma can be found in~\cite{henry2025universal}.

The use of the DSM with champernowne's constant as initial point eliminates the need for tuning parameters such as the skew value and noise threshold, which are typically required in conventional NL algorithms. Instead of using a predefined noise level to terminate the neural trace, the trace is halted when its first three decimal places match those of the input stimulus (feature value). The threshold of three decimal places was chosen arbitrarily in this study and may be adjusted in future work based on empirical performance. This is equivalent to the noise threshold of $0.001$. 

Suppose the stimulus (feature value) is $0.b_1b_2b_3b_4b_5$. Then by the lemma \ref{cmp_position}, the number $N'=b_1b_2b_3$ will occur in the decimal expansion of $c$ (Champernowne's constant) after $\alpha=3N'-(10+10^2)-1$ digits from the decimal point (Here, we are considering a threshold of three decimal places i.e, $d=3$). Thus $\alpha$ will be an upper bound for the firing time. 

Specifically, lemma \ref{cmp_position} allows us to determine firing time bound and firing rate associated with each stimulus, which serves as the key chaotic feature in the proposed variant of the Neurochaos Learning algorithm.

\section{Proposed Algorithm: AutochaosNet}
In this section, we will employ universal orbit generated by decimal shift map with Champernowne's constant to develop a hyperparameter-free Neurochaos Learning Architecture i.e, AutochaosNet. 
There are two versions of AutochaosNet based on the features extracted: 
\begin{itemize}
    \item TM AutochaosNet (Tracemean AutochaosNet)
    \item TM-FR AutochaosNet (Tracemean and Firing Rate AutochaosNet)
\end{itemize}
The proposed architecture is as follows: 

\begin{enumerate}
\item[(a)]\textbf{Input} : Suppose the input data is of size $m\times n$, where $m$ represents the number of samples and $n$ denotes the number of raw input features (or attributes). Let $$\{(x_{11},x_{12},\ldots,x_{1n}),(x_{21},x_{22},\ldots,x_{2n}),\ldots,(x_{m1},x_{m2},\ldots,x_{mn})\}$$ 
be the entire dataset. After suitable normalisation, every data instance can undergo feature transformation sequentially. Min-max normalisation is utilised here.The normalised data can be represented as : $$\{(z_{11},z_{12},\ldots,z_{1n}),(z_{21},z_{22},\ldots,z_{2n}),\ldots,(z_{m1},z_{m2},\ldots,z_{mn})\}$$
where $z_{ij}=\frac{x_{ij}-min(\{x_{ij}:1\leq i \leq m\})}{max(\{x_{ij}:1\leq i \leq m\})-min(\{x_{ij}:1\leq i \leq m\})}$ for $1\leq j \leq n$.\\
    
\item[(b)] \textbf{Feature Transformation}: Each feature attribute $x_{ij}$ of the data instance has to be transformed into a chaos-based feature.

In both versions of AutochaosNet, the firing time bound value corresponding to each $x_{ij}$ is calculated using the formula defined in Lemma \ref{cmp_position} within three decimal places. Let $T$ be the firing time bound corresponding to $x_{(ij)}$ and the Champernowne's constant (truncated till 500), $$c=0.1234567891011\ldots498499$$ be the initial point. Then the neural trace corresponding to $x_{ij}$ is $$\tau=\{c,f(c),f^{(2)}(c),\ldots,f^{(T-1)}(c)\}$$ 

\item[(c)] \textbf{Feature Extraction}: In TM AutochaosNet, corresponding to each $x_{ij}$, mean of the neural trace $\tau$ is calculated. Suppose if $t_{ij}$ is the mean of the neural trace corresponding to $x_{ij}$. Then the extracted feature set will be : \\
$$\{(t_{11},t_{12},\ldots,t_{1n}),(t_{21},t_{22},\ldots,t_{2n}),\ldots,(t_{m1},t_{m2},\ldots,t_{mn})\}$$

In TM-FR AutochaosNet, corresponding to each $x_{ij}$, mean of the neural trace $\tau$ and its firing rate are calculated. Firing rate corresponding to each $x_{ij}$ is defined as fraction of time for which the neural trace $\tau$ exceeds the value $0.5$ to recognise the stimulus.

\item[(d)] \textbf{Classification}: Datasets with the transformed features can be classified using cosine similarity. First, compute the mean representation vector of each class. 
Suppose the $m$ data instances belong to $k$ classes. Let  
$$\{(x_{l_{1}1},x_{l_{1}2},\ldots,x_{l_{1}n}),(x_{l_{2}1},x_{l_{2}2},\ldots,x_{l_{2}n}),\ldots,(x_{l_{r}1},x_{l_{r}2},\ldots,x_{l_{r}n})\}$$ be the $r$ samples in class $l$.
After normalisation, the extracted data be:
$$\{(f_{l_{1}1},f_{l_{1}2},\ldots,f_{l_{1}n}),(f_{l_{2}1},f_{l_{2}2},\ldots,f_{l_{2}n}),\ldots,(f_{l_{r}1},f_{l_{r}2},\ldots,f_{l_{r}p})\}$$
In TM AutochaosNet, the value of $p$ will be $n$ itself, since there is only one feature. But in TM-FR AutochaosNet, the value of index $p$ will be $2n$.
Then the mean representation vector corresponding to the class $l$ can be defined as:
$$M^{(l)}=\left(\frac{\sum\limits_{i=l_{1}}^{l_{r}}f_{i1}}{r},\frac{\sum\limits_{i=l_{1}}^{l_{r}}f_{i2}}{r},\ldots,\frac{\sum\limits_{i=l_{1}}^{l_{r}}f_{ip}}{r}\right) $$
In order to classify a particular data instance $$X_{i}=(x_{i1},x_{i2},\ldots,x_{in}),$$ calculate the cosine similarity of extracted feature vector of the data instance $F_{i}=(f_{i1},f_{i2},\ldots,f_{ip})$ with the mean representation vectors of each class $\{M^{(1)},M^{(2)},\ldots,M^{(k)}\}$. Cosine similarity is defined as follows:\\
$$\cos{\theta}=\frac{M^{(j)} \cdot F_{i}}{||M^{(j)}||~~||F_{i}||},$$ where $j=1,2,\ldots, k$.
The data instance will belong to the class with least value. 

\item[(e)] \textbf{Output} : The output refers to the class to which each test data instance is assigned. AutochaosNet gives output based on mean representation vector of each class .
\end{enumerate}

\begin{figure*}[htbp]
    \centering
    \includegraphics[width=18cm]{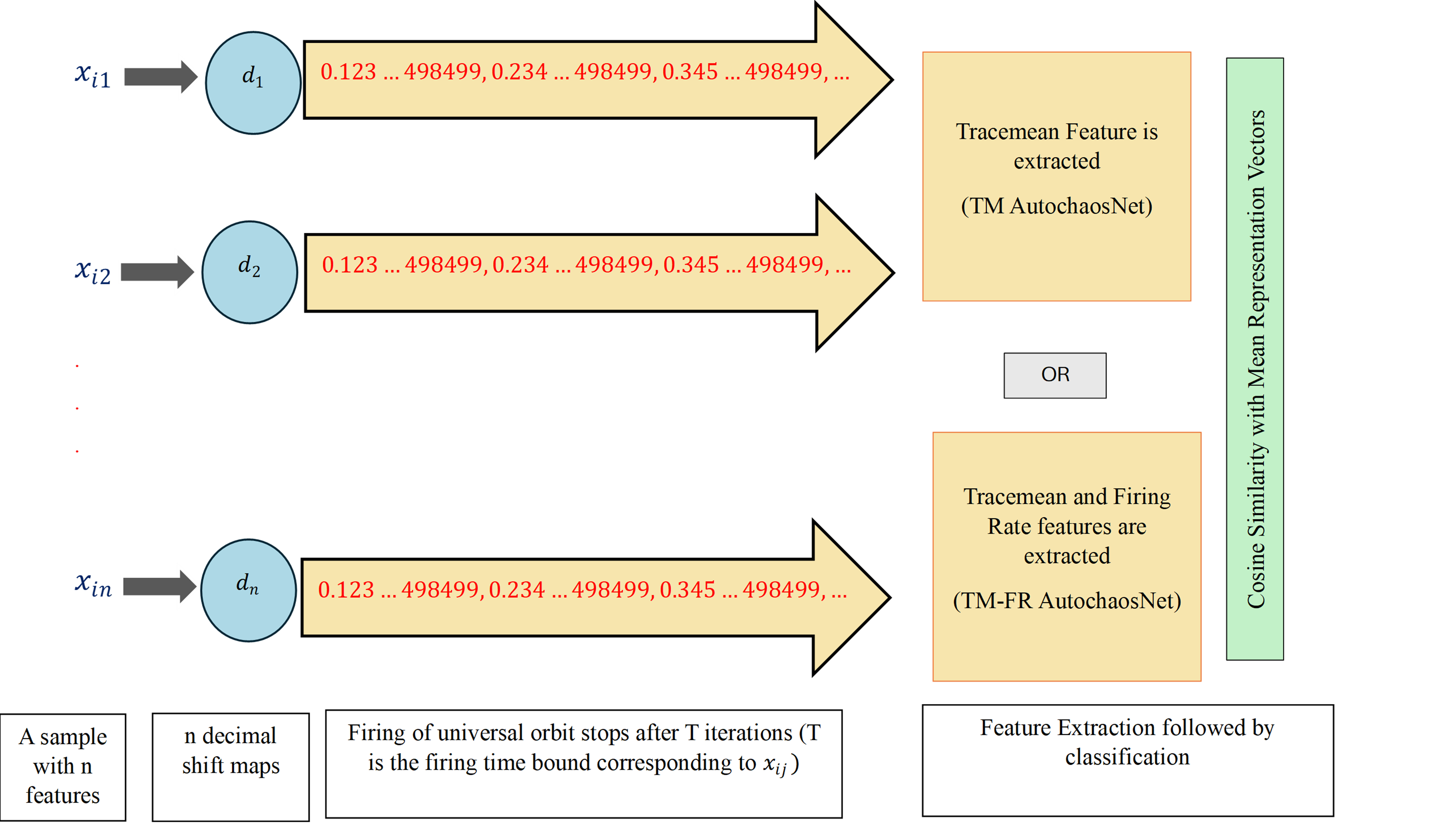}
    \caption{Architecture of AutochaosNet algorithm for feature extraction and classification.}
    \label{fig:Autochaosnet}
\end{figure*}

The proposed algorithm is illustrated in Figure \ref{fig:Autochaosnet}.

\subsection{Results and Analaysis}
This section presents a comparative analysis of the proposed hyperparameter free algorithms : TM AutochaosNet and TM-FR AutochaosNet against chaosNet using F1 Score as evaluation metric. ChaosNet algorithm utilises four chaotic features and three hyperparameters whereas the proposed algorithms utilises maximum two chaotic features and operate without any hyperparameters. The table \ref{autochaosnet f1 score} reports the F1 Scores of AutochaosNet aand chaosNet algorithms across ten distinct datasets. The F1 Scores with the highest values are highlighted in bold. Notably, despite the absence of hyperparameters, the performance of AutochaosNet algorithm remains comparable to that of chaosNet algorithm.

\begin{table}[h!]
    \centering
     \caption{\textbf{Comparison of F1 Scores for AutochaosNet with ChaosNet.}}
    \begin{tabular}{lccc}
        \toprule
        Dataset & TM-FR AutochaosNet & TM AutochaosNet & ChaosNet \\
        \midrule
        Iris & 0.868 & 0.838 & \textbf{1.000} \\
        Haberman & 0.537 & 0.516 & \textbf{0.560} \\
        Seeds & \textbf{0.878} & \textbf{0.878} & 0.845 \\
        Statlog &\textbf{ 0.755} & \textbf{0.755} & 0.738 \\
        Breast Cancer & 0.784 & 0.717 & \textbf{0.927} \\
        Bank & 0.821 & 0.806 & \textbf{0.845} \\
        Ionosphere & 0.823 & 0.813 & \textbf{0.860} \\
        Wine & 0.858 & 0.824 & \textbf{0.976} \\
        Sonar & \textbf{0.785} & 0.760 & 0.643 \\
        Penguin & 0.907 & 0.885 & \textbf{0.964} \\
        \bottomrule
    \end{tabular}
   
    \label{autochaosnet f1 score}
\end{table}
\subsubsection{Computational Complexity}
Beyond achieving comparable F1 Scores, a key advantage of AutochaosNet Algorithm is its ablity to classify datasets in a shorter time compared to chaosNet. To compare the time complexity of these algorithms, four datasets namely - Iris, Seeds, Statlog and Sonar were selected for evaluation. Each algorithm was executed over 50 iterations and the average value of elapsed time over 50 iterations was computed. The mean elapsed time over 50 iterations for TM-AutochaosNet, TM- FR AutochaosNet and ChaosNet is indicated in the table \ref{autochaosnet_time}.
\begin{table*}[h!]
    \centering
    \caption{\textbf{Comparison of Macro F1 Scores, Mean and Standard Deviation of Elapsed Time over 50 iterations for different models.}}
    \setlength{\tabcolsep}{5pt} 
    \renewcommand{\arraystretch}{1.2} 
  \resizebox{\textwidth}{!}{  \begin{tabular}{lcccccccc}
        \toprule
        \multirow{2}{*}{Dataset} & \multicolumn{3}{c}{TM-FR AutochaosNet} & \multicolumn{3}{c}{TM AutochaosNet} & \multicolumn{2}{c}{ChaosNet (Single Iteration)} \\
        \cmidrule(lr){2-4} \cmidrule(lr){5-7} \cmidrule(lr){8-9}
        & F1 Score & Mean Time (s) & Standard Deviation & F1 Score & Mean Time (s) & Standard Deviation & F1 Score & Mean Time (s)  \\
        \midrule
        \textit{Iris}     & $0.868$ & $0.741$  & $0.123$  & $0.838$ & $0.596$  & $0.018$  & $1$     & $52.15$    \\
        \textit{Seeds}    & $0.878$ & $1.577$  & $0.044$   & $0.878$ & $4.99$   & $0.475$    & $0.845$ & $7310.19$  \\
       \textit{Statlog}  & $0.755$ & $2.384$  & $0.241$  & $0.755$ & $2.531$  & $0.279$    & $0.738$ & $19750$    \\
        \textit{Sonar}    & $0.785$ & $14.506$ & $0.858$   & $0.76$  & $14.702$ & $0.584$   & $0.643$ & $10686.01$  \\
        \bottomrule
    \end{tabular}}
    
    \label{autochaosnet_time}
\end{table*}

\section{Conclusion}
In this study, we introduced two hyperparameter-free variants of the Neurochaos Learning algorithm—TM AutochaosNet and TM-FR AutochaosNet—which leverage a universal chaotic orbit generated using the Decimal Shift Map with  Champernowne's constant (truncated till 500) as the initial point. By eliminating the need for hyperparameter tuning and relying on a simplified set of chaotic features, these models significantly reduce computational complexity while preserving classification accuracy.

Extensive evaluations on multiple benchmark datasets demonstrate that the proposed AutochaosNet algorithms deliver competitive performance compared to ChaosNet. Notably, AutochaosNet achieves this while using fewer features and no hyperparameters, offering substantial gains in efficiency.
The ability of AutochaosNet to generalise across datasets without hyperparameter tuning makes it a scalable and practical solution for real-world classification problems.

In future work, we aim to identify sets of points that generate universal orbits under various chaotic maps. These truncated orbits will be used as starting triggers within the Neurochaos Learning framework, with the goal of further enhancing classification accuracy while preserving the algorithm’s simplicity and parameter-free nature.

\section*{Acknowledgements}
NN would like to acknowledge the financial support of National Board of Higher Mathematics (NBHM), Dept. of Atomic Energy, Govt. of India (Grant No. 02011/8/2025/NBHM(R.P)/R\&D II/1259).

\section*{Code Availability}
The implementation of our proposed method is available at: https://github.com/akhilahenry98/AutochaosNet.git  
\begin{appendices}
\section*{Appendix}

\section{Dataset Description}\label{Dataset}
This section presents a summary of the ten datasets employed in this study, detailing the number of samples and the number of classes associated with each dataset.
\begin{table*}[htbp]
    \caption{\textbf{Number of features and samples in each dataset used in this study.}}
    \label{data}
    \centering
    \resizebox{0.7\textwidth}{!}{
    \begin{tabular}{lccc}
 \toprule
Dataset &  Number of Features &  Number of Classes  & Number of Samples\\
 \midrule
\textit{Iris} & $4$  &  $3$  & $150$\\
\textit{Haberman's Survival}  &  $3$  &  $2$  &  $306$\\
\textit{Seeds}  &  $7$  &  $3$  &  $210$\\
\textit{Statlog (Heart)}  & $13$  &  $2$  & $270$\\
\textit{Ionosphere}  &  $34$  &  $2$  &  $351$\\
\textit{Bank Note Authentication}  &  $4$  & $2$  &  $1372$\\
\textit{Breast Cancer Wisconsin} &  $31$  &  $2$  & $569$\\
\textit{Wine}  & $13$ &  $3$ &  $178$\\
\textit{Penguin}  & $4$  & $3$  & $342$\\
\textit{Sonar}  & $60$  & $2$  & $208$\\
 \bottomrule
    \end{tabular}}
\end{table*}
\subsection{Iris}
\textit{Iris}~\cite{iris} dataset contains information about three Iris plant varities: "{\it Iris Setosa, Iris Versicolor}, and {\it Iris Virginica}". This dataset contains $150$ data objects, each having four attributes: "sepal length, width,petal length and width". All attributes are measured in {\it cms}. The class distribution in Table~\ref{iris} is as follows: \textit{Setosa, Versicolor, Virginica}.
\begin{table}[h!]
    \centering
    \caption{\textbf{Iris : Number of Samples in each class.}}
    \begin{tabular}{lcc}
    \toprule
       Class  & Label  & Number of Samples \\
       \midrule
        Iris Setosa & $0$ &$50$ \\
        Iris Versicolor & $1$ & $50$\\
        Iris Virginica &$2$  & $50$\\
        \bottomrule
    \end{tabular}
    \label{iris}
\end{table}
\subsection{Haberman's Survival}
The dataset includes breast cancer surgical survival records from a 1958–1970 study at the University of Chicago's Billings Hospital~\cite{haberman}. It contains 306 cases, each of which is characterised by three numerical characteristics: the number of positive axillary lymph nodes found, the year of surgery (represented as the year minus 1900), and the patient's age at the time of surgery. The classification is based on whether the patient died within five years after surgery or survived for at least five years, as indicated in Table~\ref{Haberman}.
\begin{table}[h!]
    \centering
     \caption{\textbf{Haberman : Number of Samples in each class.}}
    \begin{tabular}{lcc}
       \toprule
       Class  & Label  & Number of Samples \\
       \midrule
        Less than 5 years & $0$ &$225$ \\
         Greater than or equal to 5 years& $1$ & $81$ \\
         \bottomrule
    \end{tabular}
    \label{Haberman}
\end{table}
\subsection{Seeds}
The \textit{Seeds}~\cite{statlogseeds} dataset offers measurements of the geometrical features of kernels from three distinct types of wheat - {\it Kama, Rosa} and {\it Canadian}. Seven real-valued attributes namely compactness, length, width etc. were constructed using a soft X-ray approach and the GRAINS package. The class distribution is shown in Table~\ref{Seeds}.
\begin{table}[h!]
    \centering
    \caption{\textbf{Seeds: Number of Samples in each class.}}
    \begin{tabular}{ccc}
       \toprule
       Class  & Label  & Number of Samples \\
       \midrule
        Kama &$0$  & $70$\\
         Rosa& $1$ & $70$\\
        Canadian & $2$ & $70$\\
        \bottomrule
    \end{tabular}
    \label{Seeds}
\end{table}
\subsection{Statlog (Heart)}
The \textit{Statlog}~\cite{statlogseeds} dataset provides information on the absence or presence of heart disease using $13$ attributes including resting blood pressure, chest pain type, exercise induced angina and so on. The class distribution is shown in Table~\ref{Heart}.
\begin{table}[h!]
    \centering
    \caption{\textbf{Heart: Number of Samples in each class.}}
    \begin{tabular}{ccc}
       \toprule
       Class  & Label  & Number of Samples \\
       \midrule
        Absence &$0$  & $150$\\
         Presence & $1$ & $120$\\       
        \bottomrule
    \end{tabular}
    \label{Heart}
\end{table}
\subsection{Ionosphere}
The \textit{Ionosphere}~\cite{ionosphere} dataset is a binary classification dataset with dimensionality $34$. The classes indicate the status of receiving a radar signal from the Ionosphere. The objective of this experiment is to determine the composition of the Ionosphere through the analysis of radar data.The class distribution is shown in Table~\ref{Ionosphere}.
\begin{table}[h!]
    \centering
    \caption{\textbf{Ionosphere: Number of Samples in each class.}}
    \begin{tabular}{ccc}
       \toprule
       Class  & Label  & Number of Samples \\
       \midrule
        Bad &$0$  & $126$\\
         Good & $1$ & $225$\\       
        \bottomrule
    \end{tabular}
    \label{Ionosphere}
\end{table}
\subsection{Bank Note Authentication}
\textit{Bank Note Authentication}~\cite{banknote} dataset is on differentiating Genuine and Forgery currencies. Data were retrieved from photos of authentic and counterfeit banknote-like samples.The dataset contains four attributes such as variance of Wavelet Transformed image (continuous), skewness of Wavelet Transformed image (continuous) etc. obtained from the the images using wavelet transformation. The class distribution is shown in Table~\ref{Bank}.
\begin{table}[h!]
    \centering
    \caption{\textbf{Bank: Number of Samples in each class.}}
    \begin{tabular}{ccc}
       \toprule
       Class  & Label  & Number of Samples \\
       \midrule
        Genuine &$0$  & $762$\\
         Forgery & $1$ & $610$\\       
        \bottomrule
    \end{tabular}
    \label{Bank}
\end{table}
\subsection{Breast Cancer Wisconsin}
\textit{Breast Cancer Wisconsin}~\cite{breastcancer} (Diagnostic) dataset predict if the cancer is benign or malignant. The $31$ features such as radius, perimeter, texture, smoothness etc. for each cell nucleus are derived from a digitised image of a fine needle aspirate (FNA) of a breast tumour. The class distribution is shown in Table~\ref{Cancer}.
\begin{table}[h!]
    \centering
    \caption{\textbf{Cancer: Number of Samples in each class.}}
    \begin{tabular}{ccc}
       \toprule
       Class  & Label  & Number of Samples \\
       \midrule
        Malignant &$0$  & $212$\\
         Benign & $1$ & $357$\\       
        \bottomrule
    \end{tabular}
    \label{Cancer}
\end{table}
\subsection{Wine}
The results of a chemical analysis of wines produced from three different grape varieties in the same region of Italy are available in the \textit{Wine dataset}\cite{wine}. Thirteen chemical constituent concentrations were measured for every wine sample in the study. With the labels "1, 2, and 3," the wines are divided into three classes. Table\ref{Wine} shows the distribution of these classes.
\begin{table}[h!]
    \centering
    \caption{\textbf{Wine: Number of Samples in each class.}}
    \begin{tabular}{ccc}
       \toprule
       Class  & Label  & Number of Samples \\
       \midrule
        $1$&$0$  & $59$\\
         $2$ & $1$ & $71$\\ 
         $3$ & $2$ & $48$\\ 
        \bottomrule
    \end{tabular}
    \label{Wine}
\end{table}
\subsection{Palmer Penguins Dataset}
The \textit{Palmer Archipelago Penguin} data~\cite{penguins} is a dataset that includes size measurements for three species of penguins. Data was gathered from three islands in the Palmer Archipelago, Antarctica, comprising information on $344$ penguins. There are three penguin species in the dataset: {\it Adelie, Gentoo}, and {\it Chinstrap}. The class distribution is shown in Table~\ref{Penguin}.
\begin{table}[h!]
    \centering
    \caption{\textbf{Penguin: Number of Samples in each class.}}
    \begin{tabular}{ccc}
       \toprule
       Class  & Label  & Number of Samples \\
       \midrule
        Adelie Penguin &$0$  & $151$\\
        Chinstrap Penguin & $1$ & $68$\\ 
         Gentoo Penguin & $2$ & $123$\\ 
        \bottomrule
    \end{tabular}
    \label{Penguin}
\end{table}
\subsection{Sonar}
The \textit{Sonar} dataset~\cite{sonar} contains $97$ patterns from rocks under similar conditions and $111$ patterns from sonar signals bounced off a metal cylinder at various angles and conditions. The $60$ numbers in the pattern range from $0.0$ to $1.0$. The label for each record indicates whether the item is a mine (represented by "M") or a rock (represented by "R"). Table~\ref{Sonar} displays the distribution of classes.
\begin{table}[h!]
    \centering
    \caption{\textbf{Sonar: Number of Samples in each class.}}
    \begin{tabular}{ccc}
       \toprule
       Class  & Label  & Number of Samples \\
       \midrule
        Mine &$0$  & $111$\\
        Rock & $1$ & $97$\\  
        \bottomrule
    \end{tabular}
    \label{Sonar}
\end{table}

\end{appendices}

\bibliographystyle{unsrt}  
\bibliography{main}

\end{document}